\RequirePackage{amsmath, amssymb}
\documentclass[runningheads]{llncs}
\usepackage{booktabs}
\usepackage{xcolor}
\usepackage{mathtools}
\usepackage{hyperref}
\usepackage[capitalize]{cleveref}
\usepackage{svg}
\usepackage{graphicx}
\usepackage{tikz}
\usepackage{multirow}
\usepackage{array}
\usepackage{arydshln}
\usepackage{makecell}
\usepackage{xspace}

\usepackage{ifthen}
\newboolean{usebiblatex}
\setboolean{usebiblatex}{false}

\ifthenelse{\boolean{usebiblatex}}{
\usepackage[hyperref=true]{biblatex}
\DeclareCiteCommand{\cite}
  {}
  {\printtext[bibhyperref]{[\thefield{entrykey}]}} % entrykey is the citation key
  {}
  {}
\addbibresource{ref.bib}
}{}

\newcommand{\noindentbf}[1]{\noindent \textbf{#1}}
\usepackage[T1]{fontenc}
\usepackage{graphicx}
\begin{document}
% \title{Unveiling Anatomical Variations: Interpretable Adaptation on a Probabilistic Manifold for Cross-Domain Medical Image Segmentation}
\title{RemInD: Remembering Anatomical Variations for Interpretable Domain Adaptive Medical Image Segmentation}
\titlerunning{Remembering Anatomical Variations for Interpretable Domain Adaptation}
% If the paper title is too long for the running head, you can set
% an abbreviated paper title here
%
\author{Xin Wang\inst{1} \and Yin Guo\inst{2} \and Kaiyu Zhang\inst{2} \and Niranjan Balu\inst{3} \and Mahmud Mossa-Basha\inst{3} \and Linda Shapiro\inst{4} \and Chun Yuan\inst{3,5}}
\authorrunning{X. Wang et al.}
% First names are abbreviated in the running head.
% If there are more than two authors, 'et al.' is used.
%
\institute{
Department of Electrical and Computer Engineering, University of Washington, Seattle, United States \and Department of Bioengineering, University of Washington, Seattle, United States \and Vascular Imaging Lab, Department of Radiology, University of Washington, Seattle, United States \and Paul G. Allen School of Computer Science and Engineering, University of Washington, Seattle, United States \and Department of Radiology and Imaging Sciences, University of Utah, Salt Lake City, United States}
\maketitle              % typeset the header of the contribution

\newcommand{\ours}{RemInD\xspace}

\begin{abstract}
This work presents a novel Bayesian framework for unsupervised domain adaptation (UDA) in medical image segmentation. 
While prior works have explored this clinically significant task using various strategies of domain alignment, they often lack an explicit and explainable mechanism to ensure that target image features capture meaningful structural information. Besides, these methods are prone to the curse of dimensionality, inevitably leading to challenges in interpretability and computational efficiency. To address these limitations, 
% we argue that previous works have all missed a crucial prior knowledge for medical images: the tissue to segment has typical shapes. Incorporating such prior can introduce a strong inductive bias beneficial for adaptation. To this end, 
we propose \ours, a framework inspired by \emph{human adaptation}. \ours learns a domain-agnostic latent manifold, characterized by several anchors, to 
% embody all possible
memorize
anatomical variations. By mapping images onto this manifold as weighted anchor averages, our approach ensures realistic and reliable predictions. This design mirrors how humans develop representative components to understand images and then retrieve component combinations from memory to guide segmentation. 
Notably, model prediction is determined by two explainable factors: a low-dimensional anchor weight vector, and a spatial deformation. This design facilitates computationally efficient and geometry-adherent adaptation by aligning weight vectors between domains on a probability simplex.
% Particularly, a few anchors whose weighted average composed with a spatial diffeomorphism describe anatomical variation for each image. 
% Cross-domain alignment can then be performed in the low-dimensional probability simplex of anchor weights using any established strategies, significantly reducing computational costs and enhancing alignment efficiency. 
% We further propose a novel Bayesian model to implement the framework through variational inference, therefore the proposed method is both theoretically grounded and empirically reasonable. 
Experiments on two public datasets, encompassing cardiac and abdominal imaging, demonstrate the superiority of \ours, which achieves state-of-the-art performance using a single alignment approach, outperforming existing methods that often rely on multiple complex alignment strategies.
% Besides, even without any alignment strategy, our method still surpasses most of previous works, indicating effectiveness of the inductive bias from framework design.

\end{abstract}

\newcommand{\red}[1]{\textcolor{red}{#1}}
\newcommand{\myeq}[1]{%
  \begin{equation}
  \begin{aligned}
    #1
  \end{aligned}
  \end{equation}
}
\renewcommand{\mid}{\mathclose{}|\mathopen{}}
\newcommand{\x}{\boldsymbol{x}}
\newcommand{\X}{\boldsymbol{X}}
\newcommand{\Y}{\boldsymbol{Y}}
\renewcommand{\xi}{\x_i}
\newcommand{\xis}{\x_i^s}
\newcommand{\xit}{\x_i^t}
\newcommand{\y}{\boldsymbol{y}}
\newcommand{\yi}{\y_i}
\newcommand{\yis}{\y_i^s}
\newcommand{\R}{\mathbb{R}}
\newcommand{\RD}{\R^D}
\newcommand{\RDK}{\R^{D\times K}}
\newcommand{\RK}{\R^{K}}
\newcommand{\dsdef}{(\xis,\yis)_{i=1}^{N_s}}
\newcommand{\dtdef}{(\xit)_{i=1}^{N_t}}
\renewcommand{\S}{\boldsymbol{S}}
\newcommand{\Z}{\boldsymbol{Z}}
\newcommand{\Zs}{(\Z_i)_{i=1}^M}
\newcommand{\w}{\boldsymbol{w}}
\newcommand{\W}{\boldsymbol{W}}
\newcommand{\gm}{\widetilde{\w}}
\newcommand{\Zavg}{\widetilde{\Z}}
\newcommand{\st}{\boldsymbol{\phi}}
\newcommand{\fseg}{f_{\text{seg}}}
\newcommand{\fimg}{f_{\text{img}}}
\newcommand{\ie}{\emph{i.e.}}
\newcommand{\eg}{\emph{e.g.}}
\newcommand{\T}{\top}
\newcommand{\omg}{\mathrm{\Omega}}
\newcommand{\dlt}{\mathrm{\Delta}}
\newcommand{\E}{\mathbb{E}}
\renewcommand{\v}{\boldsymbol{v}}
\newcommand{\z}{\boldsymbol{z}}
\newcommand{\s}{\boldsymbol{s}}
\newcommand{\defas}{\coloneqq}
\newcommand{\post}{q(\w,\z,\v,\s\mid\x,\y)}
\newcommand{\prior}{p(\x,\y,\w,\z,\v,\s)}
\newcommand{\postnoy}{q(\w,\z,\v,\s\mid\x)}
\newcommand{\qs}{q(\s\mid\x)}
\newcommand{\qw}{q(\w\mid\x)}
\newcommand{\qz}{q(\z\mid\w)}
\newcommand{\qv}{q(\v\mid\x,\z)}
\newcommand{\pw}{p(\w)}
\newcommand{\ps}{p(\s)}
\newcommand{\pv}{p(\v)}
\newcommand{\pz}{p(\z\mid\w)}
\newcommand{\px}{p(\x\mid\z,\v,\s)}
\newcommand{\py}{p(\y\mid\z,\v)}
\newcommand{\Lxy}{\text{ELBO}}
\renewcommand{\L}{\mathcal{L}}
\newcommand{\kl}[2]{D_{\text{KL}}\left[#1\ \|\ #2\right]}
\newcommand{\Lvel}{\mathcal{L}_{\text{vel}}}
\newcommand{\Latlas}{\mathcal{L}_{\text{atlas}}}
\newcommand{\Lanchor}{\mathcal{L}_{\text{anchor}}}
\newcommand{\Lrecon}{\mathcal{L}_{\text{recon}}}
\newcommand{\Lseg}{\mathcal{L}_{\text{seg}}}
\newcommand{\N}{\mathcal{N}}
\newcommand{\mubold}{\boldsymbol{\mu}}
\newcommand{\sgmbold}{\boldsymbol{\Sigma}}
\newcommand{\bs}{B}
\newcommand{\Lalign}{\mathcal{L}_{\text{align}}}
\newcommand{\OT}{\text{OT}}
\newcommand{\e}{\varepsilon}
\newcommand{\qsgm}{F_s}
\newcommand{\qtgm}{F_t}
\renewcommand{\d}{\text{d}}
\newcommand{\dfr}{D_{\text{FR}}}
\newcommand{\Lgm}{\mathcal{L}_{\text{geo}}}
\newcommand{\lmd}{\boldsymbol{\lambda}}
\newcommand{\transplan}{\zeta}
\newcommand{\stilde}{\widetilde{\s}}
\renewcommand{\c}{\boldsymbol{c}}
\renewcommand{\circ}{\,\raisebox{1pt}{\tikz \draw[line width=0.6pt] circle(1.2pt);}\,}

% \newcolumntype{C}[1]{>{\centering\let\newline\\\arraybackslash\hspace{0pt}}m{#1}}
\renewcommand{\pm}{/}
\newcommand{\ms}[2]{#1\scalebox{1}{$\pm$}#2}
\newcommand{\msb}[2]{\textbf{#1}\scalebox{1}{$\pm$}#2}

\section{Introduction}
Creating dense annotations for deep medical image segmentation models is labor-intensive. Unsupervised domain adaptation (UDA) addresses this challenge by utilizing a labeled source dataset to improve performance on an unlabeled target domain with differing imaging patterns \cite{DA_survey}.
% key idea of DA
Its rationale lies in the shared task-relevant anatomical information between the source and target datasets.
Numerous studies have attempted to exploit this invariance through domain alignment.
% GAN-based
For example, adversarial or semi-supervised approaches \cite{ADVENT,CyCMIS,DARUNet,MAPSeg} align domains implicitly through discriminators or pseudo-labels, which, however, prioritize domain consistency over structural correctness, and thus risk the loss of critical anatomical details.
% introduced discriminators to ensure that extracted features or predicted segmentations appear indistinguishable across domains. Such implicit alignment, however, cannot guarantee that the learned features preserve critical anatomical details needed for accurate segmentation, as it focuses more on domain similarity than anatomical correctness.
% VAE, transport based
Meanwhile, previous variational or optimal transport works \cite{VarDA,VAMCEI,ot_uda}, while effective at aligning global feature distributions, are computationally expensive and may overlook the quality of individual features. 
% Moreover, computing distances in latent spaces is expensive, prone to estimation errors, and lacks direct interpretability \emph{w.r.t.} anatomical structures. 
% summarize weaknesses
To summarize, prior methods face two intrinsic challenges:
\begin{enumerate}
    \item Alignment in high-dimensional feature spaces requires substantial computational costs and often relies on approximations, leading to imprecise results.
    \item The absence of an explicit and explainable mechanism to control image features hinders their ability to capture meaningful anatomical information.
\end{enumerate}

% human adaptation
In contrast to these issues, humans learn from labeled examples by forming concepts of \emph{components} that encapsulate physiologically valid shapes \cite{human_vision}.
% For example in cardiac MRI, one typical structure is the myocardium surrounding the right ventricle, forming a circle-in-circle-like shape; with the change of slice location, this structure gradually varies into another typical structure, where the left ventricle appears. 
When encountering a new modality, they recall suitable component combinations, and adapt them with moderate warping to account for natural individual-level spatial distortions.
% This component-driven, memory-based, anatomy-oriented process enables humans to efficiently generalize across domains, ensuring structural consistency and adaptability to new inputs. 
% However, there remains a crucial gap between
% prior methods and human cognition. 
This component-driven, memory-based process allows humans to generalize efficiently across domains while maintaining structural consistency. 
However, a crucial gap remains between prior methods and human cognition.
% resulting in a rough understanding of anatomical structures.
% that lacks the specificity required for precise segmentation.

\begin{figure}[t]
\includegraphics[width=\textwidth]{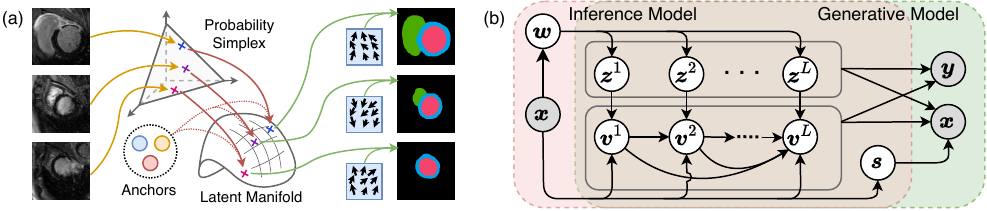}
\caption{The proposed framework \ours. (a) Based on inferred anchor weights (yellow arrows), images are mapped onto a latent manifold (red arrows), and segmentations are warped by spatial transformations to produce final predictions (green arrows). (b) Inference (pink) and generative (green) models (\cref{sec:infer}), with observations shaded.} 
\label{fig:overview}
\end{figure}

To this end, we make a radical departure from previous works by introducing \ours, a novel Bayesian framework that \emph{mimics human adaptation}. As shown in \cref{fig:overview}(a), 
\ours learns a domain-agnostic latent manifold (akin to human memory), characterized by a small set of anchors (akin to components). Each input image corresponds to a vector of anchor weights, which serves as a shape blending mechanism that enables a controlled yet flexible construction of diverse anatomical structures.
The low-dimensional weight vectors are further aligned across domains, offering a significantly more efficient and explainable alternative to traditional latent feature alignment used in previous works. 
Additionally, the segmentation predictions based on anchor weights are composed with spatial transformations to account for natural shape deformations, ensuring adaptability to diverse images.
% Consequently, the proposed method can produce realistic segmentations. 
% predictions for target images are better aligned with source-domain labels, enhancing segmentation precision.
% Given an input source or target image, the model maps it onto the manifold by predicting a set of anchor weights and a spatial diffeomorphism, analogous to how humans match a familiar shape to new observations while accounting for meaningful anatomical variations. The mapped point on the manifold is then decoded to produce the final segmentation, ensuring both anatomical precision and adaptability across diverse images. 
The contributions of our work are as follows:
\begin{enumerate}
    \item We propose \ours, a novel Bayesian UDA framework for medical image segmentation, which emulates human cognition by mapping images onto a structured manifold designed to capture the full range of anatomical variations through representative components (anchors). 
    \item \ours enables geometrically faithful adaptation within a low-dimensional probability simplex, substantially reducing computational costs while enhancing alignment efficiency and interpretability. 
    % To the best of our knowledge, this is the first deep UDA segmentation work with low-dimensional adaptation.
    \item We demonstrate the superiority of \ours over existing state-of-the-art methods, which often rely on multiple complex alignment strategies.
\end{enumerate}

\section{Methodology}

Suppose the source dataset $\dsdef$ and the target dataset $\dtdef$ consist of independent samples from the density $p_s(\X, \Y)$ and the marginal $p_t(\X)$ of $p_t(\X, \Y)$, respectively, where $\X:\RD\supset\omg\rightarrow \R$ represents the image and $\Y:\omg\rightarrow \{1,\cdots,K\}$ represents the label, with $D$ the number of pixels, and $K$ the number of classes.
We assume that the information in $\X$ can be disentangled into a content representation for segmentation and a style representation $\S$ for image appearance. Inspired by atlas-based segmentation \cite{atlas_survey}, we further disentangle the content into two independent variables: an atlas representation $\Z$ and a spatial transformation $\st: \omg\rightarrow\omg$, such that $\X\circ\st$ is registered to $\Z$. 
Considering that a single atlas is not topologically diverse, we assume the atlas is conditioned on a vector $\W$, \ie, $\Z\sim p(\Z\mid\W)$. 
% Considering the topological diversity of $\Y$, a single atlas with various $\st$ could be insufficient to fit all possible segmentations. 
% Therefore, we assume that the atlas is conditioned on a low-dimensional vector $\W$, \ie, $\Z\sim p(\Z\mid\W)$. 
% Then, $\X,\Y$ are generated as
% $\X
% =\fimg(\Z,\S)\circ\st^{-1},
% \Y
% =\fseg(\Z)\circ\st^{-1},$
% where $\st$ captures fine details of structural differences between $\Z$ and $\X$.
% Therefore, we propose modeling the generation of $(\X,\Y)$ through the following process: 
% 1) A small group of anchor distributions represents several typical shapes of the segmentation, 2) the atlas $\Z$ follows a weighted average distribution of the anchors, where the weight $\w$ allows for blending distinct anchor shapes together, 3) a smooth spatial transformation $\st$ captures fine details of structural difference between $\Z$ and $\X$, and 4) $\X,\Y$ are generated by two transformation-equivariant functions $\fimg$ and $\fseg$, respectively, \ie,
% \myeq{
% \X=\fimg(\Z\circ\st,\S)=\fimg(\Z,\S)\circ\st,\quad\Y=\fseg(\Z\circ\st)=\fseg(\Z)\circ\st.
% }
This model allows for a flexible representation of anatomy in observations through controllable factors,
\ie, a low-dimensional vector $\W$ 
% on the probability simplex $\dlt = \left\{ 
% \w \in \mathbb{R}^M \, \middle| \, 
% \w \succeq\mathbf{0}, 
% \mathbf{1}^\T \w = 1
% \right\}$ 
and a diffeomorphism $\st$.

\subsection{Bayesian Inference of Latent Variables}
\label{sec:infer}
Let $(\x,\y)$ be an observation of $(\X,\Y)$. We propose a novel variational Bayesian framework, \ours, to infer the corresponding latent representations: $\w$, $\z$, $\s$, and the stationary velocity field $\v$ parameterizing $\st$ \cite{velocity}. Specifically, we make two independence assumptions: 1) $\x$ captures all information about the latent variables, and 2) the style code $\s$ is conditionally independent of structure-related latent variables given $\x$. Hence, the joint distribution can be factorized as $\prior=\pw\ps\pv\pz\px\py$, and similarly, the variational posterior becomes $\post=\qs\qw\qz\qv$. Following the variational Bayes framework \cite{vae}, the evidence lower bound (ELBO) of the log-likelihood is expressed as
\myeq{
&\Lxy(\x,\y)\defas \E_{\post}\log \frac{\prior}{\post}\\
=&\underbrace{\E_{\qs\qw\qz\qv}\log \px}_{\Lrecon(\x)}+\underbrace{\E_{\qw\qz\qv}\log\py}_{\Lseg(\x,\y)}\\
&- \kl{\qs}{\ps}-\kl{\qw}{\pw} \\
&- \underbrace{\E_{\qw}\kl{\qz}{\pz}}_{\Latlas(\x)} - \underbrace{\E_{\qw\qz}\kl{\qv}{\pv}}_{\Lvel(\x)},
\label{eq:elbo_wrt_kls}
}
where $D_{\text{KL}}$ denotes the Kullback-Leibler (KL) divergence. Intuitively, $\Lrecon$ and $\Lseg$ correspond to image reconstruction and segmentation, while the other KL terms serve as regularization for the latent variables. For unlabeled images, a similar derivation applies, with the only difference being the omission of $\Lseg$. 

To enhance the expressiveness of $\z$ and $\v$, we decompose them hierarchically \cite{nvae} as $\z=(\z^l)_{l=1}^{L}$ and $\v=(\v^l)_{l=1}^{L}$, similar to \cite{BInGo,mri_cal_seg}, where a larger $l$ correspond to a finer spatial resolution. This construction expresses complex information in $\z$ and $\st$ by simpler components to facilitate learning. The final spatial transformation can then be calculated as $\st=\st^1\circ\cdots\circ\st^{L}$, where $\frac{\partial}{\partial t}\st^l(a,t)=\v^l(\st^l(a,t)), \forall a\in\omg,t\in[0,1]$. 
We further assume 1) different levels of $\z^l$ are independent given $\w$, 2) $\v^l$ can be inferred directly from $\x$ and $\z^l$ at the same level, and 3) the prior $p(\v^l\mid \v^{<l})=p(\v^l)$, where $<l$ denotes levels below $l$. Therefore, the KL terms $\Latlas$ and $\Lvel$ can be simplified as 
\myeq{
\Latlas(\x)=&\E_{\qw}\left\{\sum_{l=1}^{L}\kl{q(\z^l\mid\w)}{p(\z^l\mid \w)}\right\},\\
\Lvel(\x)=&\E_{\qw q(\z\mid\w)}\left[\sum_{l=1}^{L}\E_{q(\v^{<l}\mid \x,\z^{<l})}\left\{\kl{q(\v^l\mid\x,\z^l,\v^{<l})}{p(\v^l)}\right\}\right],
\label{Eq:levelwise_kl}
}
with $q(\v^{<1}\mid\x,\z^{<1})\defas 1$ for simplicity. Thus, the graphical model corresponding to the ELBO is illustrated in \cref{fig:overview}(b). This formulation enables level-wise inference and regularization of $\z$ and $\v$. Specifically, $\Lvel$ is calculated through the technique introduced in \cite{kl_v} to guarantee a diffeomorphic $\st$, while $\Latlas$ and $q(\z^l\mid\w)$ facilitate retrieving anatomical information akin to human visual recognition, as detailed in the next section.

\subsection{Anchor-Based Manifold Embedding for Interpretable Representation Extraction}

In \ours, the atlas representation $\z$ is inferred based on $\w$ through the posteriors $q(\z^l\mid\w)$ for the given image $\x$. To make this procedure explainable, we propose learning \emph{anchor} distributions $\{q_m(\z^l)\}_{m=1}^M$ for each level $l$, with $M$ the length of $\w$. To note, these anchors are not conditioned on $\x$. We further assume $q(\z^l\mid\w)$ to be the $\w$-weighted geometric mean \cite{geo_mean} of the anchors, \ie, $q(\z^l\mid\w) \propto\prod_{m=1}^M q_m^{w_m}(\z^l)$,
% \myeq{
% q(\z^l\mid\w) =\prod_{m=1}^M q_m^{\w_m}(\z^l) \left/ \int \prod_{m=1}^M q_m^{\w_m}(\z^l)\d\z^l \right.
% }
where $\w\in \dlt\defas\left\{ 
\w \in \mathbb{R}^M \, \middle| \, 
\w \succeq\mathbf{0}, 
\mathbf{1}^\T \w = 1
\right\}$, the ($M-1$)-dimensional probability simplex. Therefore, the effects of $\w$ can be interpreted as follows: 1) $\w$ serves as a shape blending weight, blending distinct shapes represented by the anchors to form an atlas distribution that encodes a new shape to fit $\x$; 2) This process essentially constructs a probabilistic manifold characterized by the anchors, which mimics humans retrieving suitable combinations of learned components  \cite{human_vision} to form anatomical shapes for segmentation. 
Moreover, through anchors not conditioned on $\x$, we impose a strong inductive bias, capturing global anatomical representations and enhancing domain adaptation by providing domain-agnostic information across various images. In sharp contrast, prior works extract representations directly from $\x$, which could be too flexible to ensure reasonable structures for images in the target domain.

We model $q_m(\z^l)$ as diagonal Gaussian distributions $\N(\mubold_m^l,\sgmbold_m^l)$. Consequently, $q(\z^l\mid\w)$ are also Gaussian, $\N(\mubold^l,\sgmbold^l)$, with
\myeq{
\mubold^l = \sgmbold^l \sum_{m=1}^M w_m \left(\sgmbold_m^l\right)^{-1} \mubold_m^l, \quad
\left(\sgmbold^l\right)^{-1} = \sum_{m=1}^M w_m \left(\sgmbold_m^l\right)^{-1}.
\label{eq:poe}
}
Besides, the prior distributions $p(\z^l\mid\w)$ are set as standard Gaussians $\N(\boldsymbol{0},\boldsymbol{I})$. Therefore, $\Latlas$ involves calculating KLs between $q(\z^l|\w)$ and $\N(\boldsymbol{0},\boldsymbol{I})$ for each image in a training mini-batch.
We further propose replacing $\Latlas$ with $\Lanchor$, defined as the average of KLs for each anchor, \ie 
\myeq{
\Lanchor = \sum_{l=1}^L\frac{1}{M}\sum_{m=1}^M\kl{q_m(\z^l)}{\N(\boldsymbol{0},\boldsymbol{I})}. 
}
Notably, calculating $\Lanchor$ does not depend on $\x$. The rationale for the replacement is twofold: 1) It can be proven that $``\Lanchor=0" \Leftrightarrow ``\forall m,l,\ q_m(\z^l)=\N(\boldsymbol{0},\boldsymbol{I})" \Rightarrow ``\Latlas=0"$, and thus, minimizing $\Lanchor$ effectively minimizes $\Latlas$, and 2) $\Lanchor$ is much more computationally efficient, requiring only $LM$ KL calculations, compared to $L\bs$ calculations for $\Latlas$, where $\bs$ is the batch size and typically $\bs\gg M$. Intuitively, $\Lanchor$ encourages the anchors $q_m(\z^l)$ to stay close to a fixed location (standard Gaussian), preventing small variations in $\w$ from causing excessive divergence in $q(\z^l\mid\w)$. 

\subsection{Efficient, Geometrically Faithful Domain Alignment}

A common approach in UDA involves aligning feature distributions between domains. Previous studies universally operate in high-dimensional latent spaces for this purpose, which is implicit, computationally expensive, and often requires approximations that compromise accuracy.
In contrast, \ours offers a computationally efficient and interpretable alternative: Since anchors are shared across all images, aligning atlas distributions $q(\z\mid\w)$ between domains reduces to aligning the shape blending weights $\w$ directly. 
  
To this end, we propose applying an optimal transport loss to the simplex $\dlt$. Specifically, we assume a deterministic posterior for the weights $\w$, similar to VQ-VAE \cite{VQVAE}, \ie, $q(\w\mid\x)=\delta(\w-\gm(\x))$, with $\delta$ the Dirac delta, and $\gm\in\dlt$ inferred from $\x$. Thus, $\kl{q(\w\mid\x)}{p(\w)}$ in \cref{eq:elbo_wrt_kls} vanishes as a constant by setting $p(\w)$ to the standard Dirichlet $\text{Dir}(\boldsymbol{1})$. More importantly, for source and target mini-batches, represented by $(\gm_i^s)_{i=1}^{\bs_s}$ and $(\gm_i^t)_{i=1}^{\bs_t}$ , we define the empirical distribution functions $F_s(\gm)\defas \frac{1}{\bs_s}\sum_{i=1}^{\bs_s}\boldsymbol{1}\{\gm_i^s\preceq\gm\}$, and similarly $F_t$, where $\boldsymbol{1}\{\cdot\}$ is the indicator function. The alignment loss is then defined as their Sinkhorn divergence \cite{sinkhorn}, \ie
\myeq{
\Lalign &\defas \OT_\e(\qsgm,\qtgm)-\frac12\OT_\e(\qsgm,\qsgm)-\frac12\OT_\e(\qtgm,\qtgm),\\
\text{with}&\ \OT_\e(F,F^\prime)\defas \min_{\transplan \in \prod(F, F^\prime)}\int_{\dlt\times\dlt}C \d\transplan +\e\int_{\dlt\times\dlt} \log \left(\frac{\d\transplan}{\d F\d F^\prime}\right)\d\transplan.
}
Here, $\e$ controls the strength of entropy regularization, $\transplan$ is a transport plan, $\prod(F,F^\prime)$ denotes all probabilistic couplings over $\dlt\times\dlt$ with marginals $F$ and $F^\prime$, and $C$ is a symmetric non-negative cost function.  
We impose the Fisher-Rao metric $\dfr$ on $\dlt$ to reflect its non-Euclidean statistical geometry, \ie
\myeq{
C(\gm,\gm^\prime)\defas\dfr(\gm,\gm^\prime)\defas2\arccos\left(\sum_{i=1}^M\sqrt{\widetilde{w}_i \widetilde{w}^\prime_i}\right),\ \forall \gm,\gm^\prime \in \dlt.
\label{eq:cost}
}
Therefore, $C$ measures the geodesic distances among $\w$ on a Riemannian manifold \cite{fisher-rao}, ensuring an intrinsic and geometrically faithful optimal transport.
Since $\dlt$ is low-dimensional, $\Lalign$ can be calculated efficiently with minimal computational overhead. Considering $\dfr\in[0,\pi]$, we set $\e=\pi/10$.

\subsection{Geometry Regularization and Final loss}
We propose regularizing $\gm$ of labeled source images through an additional loss
\myeq{
\Lgm\defas \sum_{i=1}^{\bs_s}\sum_{j=i+1}^{\bs_s}\left[\left(1-\frac{\dfr(\gm^s_i,\gm^s_j)}{\pi}\right)-\text{Sim}\left(\y^s_i\circ\st^s_i,\y^s_j\circ\st^s_j\right)\right]^2,
}
where $\text{Sim}$ is the Dice similarity coefficient, and the denominator $\pi$ normalizes $\dfr$ to $[0,1]$. This regularization loss offers several benefits: 1) It establishes a principled association between the distances among $\gm$ and the structural differences among images, explicitly refining geometry of the latent manifold to better capture anatomical variations, 2) It inherently translates the cross-domain alignment of $\gm$ to the alignment of segmentation semantics, improving the interpretability of $\Lalign$, 3) It accelerates training convergence by preventing $\gm$ of disparate images from collapsing into a single value, a pervasive issue in other works, \eg, mode a similar regularization due to the unaffordable computational cost of operating on high-dimensional representations.

The final loss is the negative of ELBO, plus $\Lalign$ and $\Lgm$. For the remaining terms in the ELBO: We model $p(\x\mid \z,\v,\s)$ as a Laplacian distribution, factorized for each pixel. Consequently, $-\Lrecon$ becomes the scaled L1 loss between the input image and its reconstruction, with pixelwise scales predicted alongside the reconstruction by a decoder. For the segmentation term $-\Lseg$ on the source domain, we utilize a combination of the cross-entropy loss and Dice loss, following prior works \cite{VAMCEI}.
Besides, the style code $\s$ is inferred deterministically, similar to $\w$. As a result, the KL for $q(\s\mid\x)$ in \cref{eq:elbo_wrt_kls} vanishes. 
Therefore, given source and target training batches $(\x_i^s,\y_i^s)_{i=1}^{\bs_s}$ and $(\x_i^t)_{i=1}^{\bs_t}$, the final loss, with $\lmd$ controlling term weights, is given by
\myeq{
\L =&  
-\frac{1}{\bs_s}\sum_{i=1}^{\bs_s}\Lseg(\x_i^s,\y_i^s) 
-\lambda_1\left[\frac{1}{\bs_s}\sum_{i=1}^{\bs_s}\Lrecon(\x_i^s)
    +\frac{1}{\bs_t}\sum_{i=1}^{\bs_t}\Lrecon(\x_i^t)\right]\\
&+ \lambda_2\left[\frac{1}{\bs_s}\sum_{i=1}^{\bs_s}\Lvel(\x_i^s)
    +\frac{1}{\bs_t}\sum_{i=1}^{\bs_t}\Lvel(\x_i^t)\right] + \lambda_3\Lanchor + \lambda_4\Lalign + \lambda_5\Lgm.
}

\begin{figure}[t]
\includegraphics[width=\textwidth]{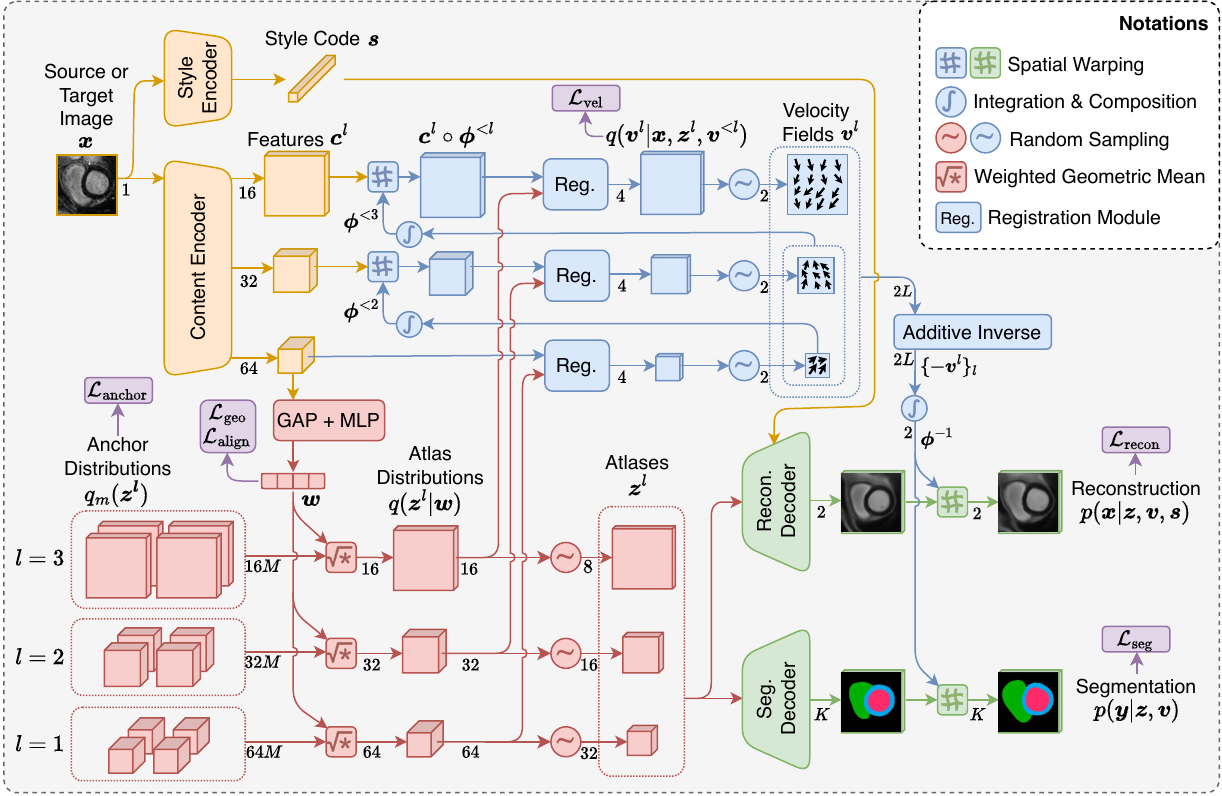}
\caption{Network architecture of \ours, illustrated with $L=3$ and $M=4$ as an example, including image feature encoding (orange), atlas inference (red), spatial transformation inference (blue), segmentation \& reconstruction decoding (green), and loss calculations (purple). The feature maps parameterizing posterior distributions contain one half of channels for the mean and the other half for the log variance. Values around arrows indicate channel numbers.} 
\label{fig:network_arch}
\end{figure}

\subsection{Network Architecture}
We design a dedicated variational autoencoder (VAE) to facilitate inference and loss calculation within \ours, as shown in \cref{fig:network_arch}. 
Given a source or target image $\x$, the content encoder extracts multilevel features, denoted as $\{\c^l\}_{l=1}^L$, and the style encoder extracts $\s$ as the style code.
For atlas inference, a multilayer perceptron (MLP) followed by a Softmax predicts the shape blending weight $\w=\gm$, based on the global average pooling (GAP) of the bottom-level feature $\c^1$. The anchor distributions $\{q_m(\z^l)\}_{m=1}^M$ are modeled using learnable parameters $\mubold^l_m,\log\sgmbold^l_m$. Therefore, for each level $l$, the atlas distribution $q(\z^l\mid\w)$ is calculated as the $\w$-weighted geometric mean of the anchors (\cref{eq:poe}), from which the atlas $\z^l$ is randomly sampled during training or inferred as the mathematical expectation during evaluation. 

To infer the velocity fields $\{\v^l\}_{l=1}^L$, similar to \cite{BInGo,bingo_arxiv}, we start at the bottom (coarsest) level $l=1$, where a registration module predicts the posterior $q(\v^1\mid \x,\z^1)$ based on the level-1 atlas distribution $q(\z^1\mid\w)$ and the feature $\c^1$. $\v^1$ is produced from the velocity posterior similar to the atlas $\z^l$. For each higher level $l>1$, the level-$l$ feature $\c^l$ is warped by $\st^{<l}\defas\st^1\circ\cdots\circ\st^{l-1}$, and the corresponding registration module predicts $q(\v^l\mid \x,\z^l,\v^{<l})$ based on the level-$l$ atlas distribution and the warped feature, where $\v^l$ is inferred similar to $\v^1$. Once all velocity fields are obtained, the final spatial transformation $\st$ and its inverse mapping $\st^{-1}$ are deterministically calculated \cite{velocity}.

For decoding, the atlases and the style code are required to provide anatomical and appearance information, respectively. Therefore, the reconstruction decoder takes both as input, while the segmentation decoder only utilizes the atlases. Outputs of the decoders are further warped by $\st^{-1}$ to produce the final reconstruction and segmentation for the image $\x$.

\section{Experiments and Results}
\subsection{Datasets}
\noindentbf{MS-CMR.} The MS-CMRSeg challenge \cite{mscmr} provides cardiac MRI images in three sequences: bSSFP, LGE and T2, with labels for the left/right ventricles (LV/RV) and myocardium (Myo). Following \cite{VarDA}, we used 35 bSSFP images as the source dataset and 45 LGE images as the target dataset, with 5/40 LGE images allocated for validation/test. All images were shuffled to be unpaired, resampled to a 0.76-mm spacing, and cropped to $192\times 192$.

\noindentbf{AMOS.} The AMOS challenge \cite{amos22} provides a multi-center, multi-disease dataset of unpaired abdominal 3D CT and MRI scans. In this study, we focused on the segmentation of liver, spleen and right/left kidneys. 
We randomly selected 25 MRI scans as the source dataset and 35 CT scans as the target dataset, with CT scans randomly split into 25/5/5 for training/validation/test. 
Axial slices were resampled to a 1.5-mm spacing and cropped to ensure a consistent field of view centered on the organs of interest, following \cite{SIFA}.

\subsection{Experimental Setups}
\noindentbf{Implementation Details.} All images were min-max normalized.
For the model architecture, we set $L=5$ and $M=6$. Moreover, the content encoder is an attention U-Net \cite{att_unet}, and the features $\{\c^l\}$ are the outputs of the attention layers. The style encoder is a Conv-LeakyReLU-Pool-Linear sequence, producing a 64-dimensional style code. The reconstruction and segmentation decoders share the same structure as the decoding part of a U-Net \cite{unet}, while adaptive instance normalizations \cite{AdaIN} are used in the reconstruction decoder to modulate feature maps with the style code. Each registration module contains four Conv-BatchNorm-LeakyReLU sequences followed by a final Conv to adjust the channel number. Experiments were conducted with PyTorch \cite{pytorch} on an NVIDIA RTX 4090 GPU. 

\noindentbf{Compared Methods and Evaluation Metrics.} We compared \ours with state-of-the-art works that utilize various adaptation methods, including VAMCEI \cite{VAMCEI}, MAPSeg \cite{MAPSeg}, DARUNet \cite{DARUNet}, VarDA \cite{VarDA}, CyCMIS \cite{CyCMIS}, and ADVENT \cite{ADVENT}. Results from an attention U-Net trained purely on the source domain are also presented as NoAdapt. For evaluation, we reported the mean and per-class Dice Similarity Coefficients (DSCs) and Average Symmetric Surface Distances (ASSDs). For CyCMIS without publicly available code, we reported the results from its publication for the overlapping dataset.

\subsection{Results}

\begin{table}[t]
\caption{Quantitative results (mean$\pm$standard deviation) of the compared methods on the MS-CMR dataset, with the best performance highlighted in bold. \#Align denotes the number of loss terms for domain alignment.
% N/A indicates methods without publicly available and functional code for reproducing results did not report the corresponding metric in their original publication.
}
\centering
\scriptsize
\label{tab:quan_mscmr}
\setlength{\tabcolsep}{0.31em} % for table column spacing
% \fontsize{7pt}{8pt}\selectfont
\begin{tabular}{lcccccccccc}
\toprule
\multirow{2}{*}{Method}& \multirow{2}{*}{\#Align} &\multicolumn{4}{c}{DSC (\%) $\uparrow$} & & \multicolumn{4}{c}{ASSD (mm) $\downarrow$}\\
\cline{3-6} \cline{8-11}
  && Mean & Myo & LV & RV && Mean & Myo & LV & RV  \\
\cline{1-11}
NoAdapt & 0 & \ms{46.2}{15} & \ms{32.8}{19} & \ms{57.2}{18} & \ms{48.6}{15}  & &\ms{18}{11} & \ms{15}{14} & \ms{16}{15} & \ms{23}{15}\\
\hdashline
ADVENT & 3 & \ms{69.7}{17} & \ms{58.1}{17} & \ms{77.8}{17} & \ms{73.3}{20} & &\ms{4.2}{12} & \ms{6.8}{29} & \ms{4.1}{6.6} & \ms{1.8}{1.0}\\
CyCMIS &11 & \ms{79.1}{7.9} & \ms{71.4}{7.3} & \ms{87.2}{7.9} & \ms{78.7}{8.5}   & &\ms{1.7}{1.5} & \ms{1.5}{1.4} & \msb{1.3}{0.9} & \ms{2.3}{2.2}\\
VarDA & 1 & \ms{79.8}{9.3} & \ms{73.0}{8.3} & \ms{88.1}{4.8} & \ms{78.5}{14.9}   & &\ms{2.6}{1.3} & \ms{1.7}{0.6} & \ms{2.6}{1.2} & \ms{3.5}{2.2}\\
DARUNet & 7 & \ms{82.0}{6.9} & \ms{75.0}{9.6} & \ms{88.4}{5.5} & \ms{82.7}{9.3}  & &\ms{1.7}{0.9} & \ms{1.3}{0.7} & \ms{2.2}{1.7} & \ms{1.6}{1.1}\\
MAPSeg & 3 & \ms{66.0}{11} & \ms{56.9}{9.7} & \ms{75.5}{13} & \ms{65.6}{14} && \ms{5.2}{10} & \ms{7.0}{27} & \ms{4.0}{7.0} & \ms{4.6}{5.0}\\
% VAMCEI & 3 & \ms{82.8}{8.5} & \ms{76.5}{8.1} & \msb{89.6}{4.3} & \ms{82.1}{13} && \ms{1.4}{0.8} & \ms{1.3}{0.5} & \ms{1.3}{1.2} & \ms{1.5}{0.9}\\
VAMCEI & 3 & \ms{82.5}{5.5} & \ms{75.8}{6.7} & \ms{88.2}{5.3} & \ms{83.5}{8.6} && \ms{1.4}{0.6} & \ms{1.1}{0.3} & \ms{1.7}{1.1} & \ms{1.5}{1.0}\\
\hdashline
\ours &1  & \msb{83.1}{5.3} & \msb{77.1}{5.8} & \msb{88.6}{4.5} & \msb{83.5}{7.5}  & &\msb{1.3}{0.6} & \msb{0.9}{0.3} & \ms{1.6}{1.1} & \msb{1.4}{0.8}\\
\bottomrule
\end{tabular}
\end{table}

\noindentbf{Quantitative Comparison.} The evaluation metrics of the compared methods on the two datasets are presented in \cref{tab:quan_mscmr} and \cref{tab:quan_amos}. Unlike the baselines, which typically employ multiple complex alignment strategies, 
\ours relies solely on a single alignment term ($\Lalign$). 
% For example, VAMCEI aligns aggregated variational posteriors and incorporates adversarial learning in the segmentation space, while DDFSeg and DARUNet utilize a considerable number of adversarial losses. 
Despite this simplicity, \ours consistently outperforms the baselines in both average DSC and ASSD. 
Notably, it achieves significant improvements for smaller, more challenging structures. For example, it improves the ASSD of myocardium by 18\%, the DSC of right kidney by 7\%, and the ASSDs of left and right kidneys by 39\% and 42\%, respectively, compared to the best-performing baselines. This superior performance can be attributed to \ours's ability to memorize anatomical structures from labeled images and adapt through shape blending weights, which allows for preserving fine-grained details. In contrast, previous methods directly align image features with spatial dimensions, which may dilute focus on less prominent structures.

\begin{figure}[t!]
\includegraphics[width=\textwidth]{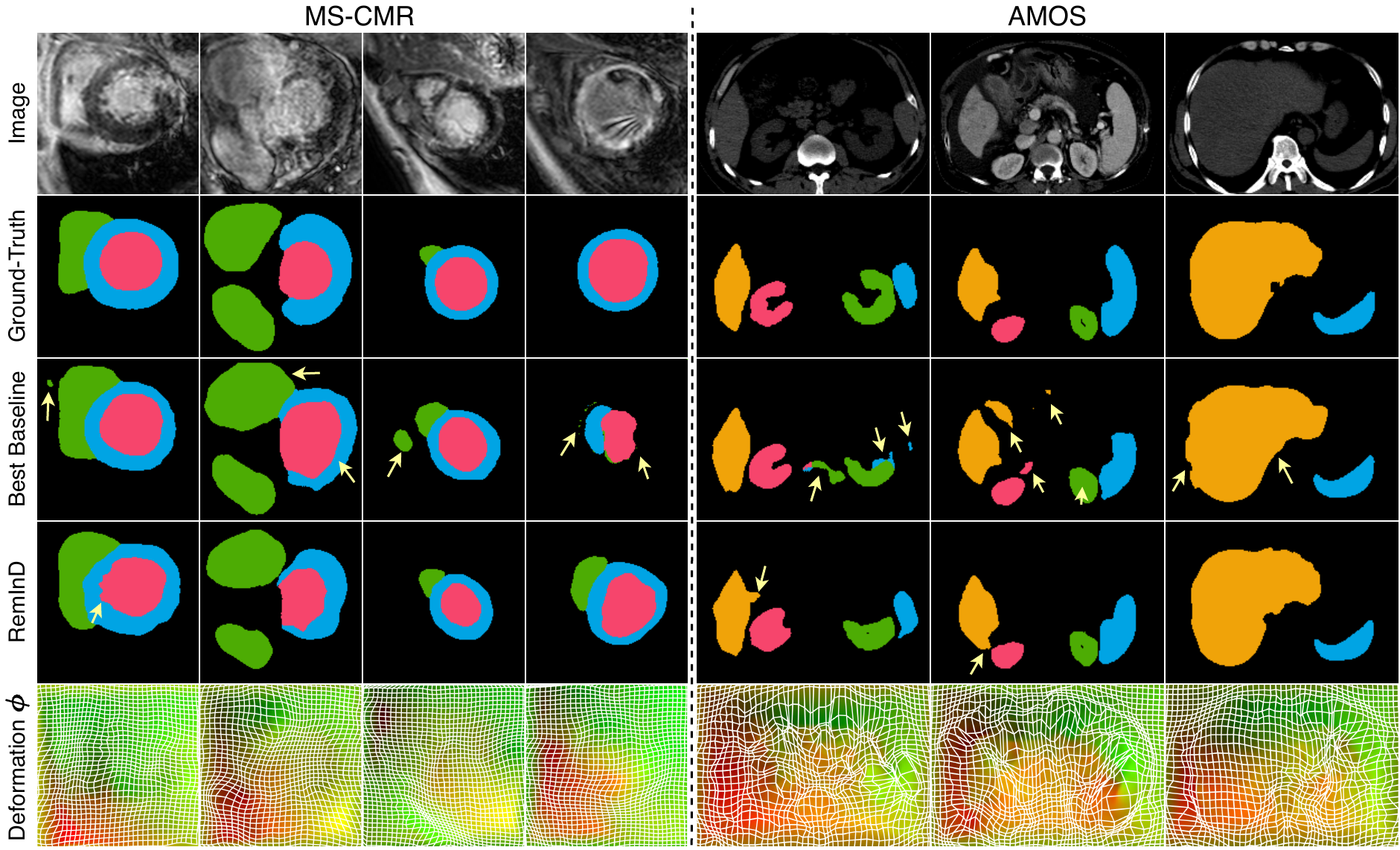}
\caption{Qualitative comparison between \ours and the best baselines. The spatial deformations $\st$ from \ours are also displayed. Yellow arrows indicate regions where one method produces inferior predictions compared to the other.} 
\label{fig:quali_compare}
\end{figure}

\noindentbf{Qualitative Comparison.} The results for example images by \ours and the best-performing baselines (VAMCEI for MS-CMR and DARUNet for AMOS) are illustrated in \cref{fig:quali_compare}. Visually, \ours generally achieves better performance, particularly in challenging cases with poor imaging quality or artifacts. Baseline methods often produce fragmented or anatomically implausible structures, such as broken ventricles or myocardium, broken or merged kidneys and spleen, and other irregularities. In contrast, \ours delivers robust predictions closely aligned with real-world labels. 
Moreover, baseline methods often rely heavily on intensity information, leading to segmentation errors. For example, as shown in the last column, some ribs and the postcava are misclassified as liver by the baseline (indicated by the two arrows), while \ours avoids these mistakes. This is notable given that the ribs appear bright and the postcava shares similar intensity with and is connected to the liver.  
This demonstrates that \ours effectively learns anatomical knowledge through the anchors and  domain alignment via shape blending weights, thereby enhancing prediction quality in the target domain. 
Additionally, the flexible displacement fields bridge the gap between individual images and the inferred atlases, capturing detailed shape variations and improving the model's capability in an interpretable manner.

\begin{table}[t]
\caption{Quantitative results (mean/standard deviation) of the compared methods on the AMOS dataset, with the best performance highlighted in bold.}
\centering
\scriptsize
\label{tab:quan_amos}
% \fontsize{0.8em}{9pt}\selectfont
\setlength{\tabcolsep}{0.14em} % for table column spacing
\begin{tabular}{cccccccccccc}
\toprule
\multirow{2}{*}{Method}&\multicolumn{5}{c}{DSC (\%) $\uparrow$} & & \multicolumn{5}{c}{ASSD (mm) $\downarrow$}\\
\cline{2-6} \cline{8-12}
  & Mean & Liver & \makecell{Left \\ Kidney} & \makecell{Right \\ Kidney} & Spleen && Mean & Liver & \makecell{Left \\ Kidney} & \makecell{Right \\ Kidney} & Spleen \\
\cline{1-12}
NoAdapt & \ms{9.4}{9.7} & \ms{29}{28} & \ms{0.2}{0.4} & \ms{4.8}{6.8} & \ms{3.2}{6.6} & &\ms{61}{19} & \ms{31}{9.1} & \ms{73}{22} & \ms{61}{31} & \ms{80}{27}\\
\hdashline
ADVENT & \ms{65.0}{6.7} & \ms{74.9}{35} & \ms{54.2}{15} & \ms{60.0}{13} & \ms{71.0}{7.4} & &\ms{5.9}{2.4} & \ms{5.3}{6.7} & \ms{8.0}{2.7} & \ms{5.1}{2.2} & \ms{5.3}{3.5}\\
VarDA & \ms{81.4}{5.3} & \ms{85.2}{7.5} & \ms{79.8}{9.9} & \ms{78.4}{7.6} & \ms{82.0}{8.4} & &\ms{4.0}{1.8} & \ms{5.3}{4.6} & \ms{4.5}{0.9} & \ms{2.7}{0.6} & \msb{3.4}{3.1}\\
DARUNet & \ms{85.8}{5.8} & \msb{91.6}{4.3} & \ms{82.4}{14} & \ms{82.1}{11} & \ms{87.1}{6.2} & &\ms{4.0}{2.4} & \ms{3.1}{2.1} & \ms{3.0}{1.9} & \ms{5.4}{3.6} & \ms{4.5}{4.7}\\
MAPSeg & \ms{85.9}{4.5} & \ms{85.1}{19} & \ms{85.3}{3.5} & \ms{82.0}{3.9} & \msb{91.0}{3.5} & &\ms{8.6}{1.8} & \ms{4.8}{7.0} & \ms{13}{9.6} & \ms{10}{6.1} & \ms{6.1}{2.8}\\
VAMCEI & \ms{84.8}{5.5} & \ms{90.3}{5.0} & \ms{82.2}{13} & \ms{80.5}{9.7} & \ms{86.2}{7.2} & &\ms{3.0}{2.0} & \msb{2.8}{2.3} & \ms{2.3}{1.4} & \ms{2.4}{1.2} & \ms{4.3}{4.4}\\
\hdashline
\ours & \msb{87.0}{2.0} & \ms{85.8}{7.2} & \msb{87.8}{2.7} & \msb{88.1}{2.6} & \ms{86.3}{5.3} & &\msb{2.8}{1.5} & \ms{4.7}{2.9} & \msb{1.4}{0.3} & \msb{1.4}{0.2} & \ms{3.7}{2.8}\\

\bottomrule
\end{tabular}
\end{table}

\begin{figure}[ht!]
\includegraphics[width=\textwidth]{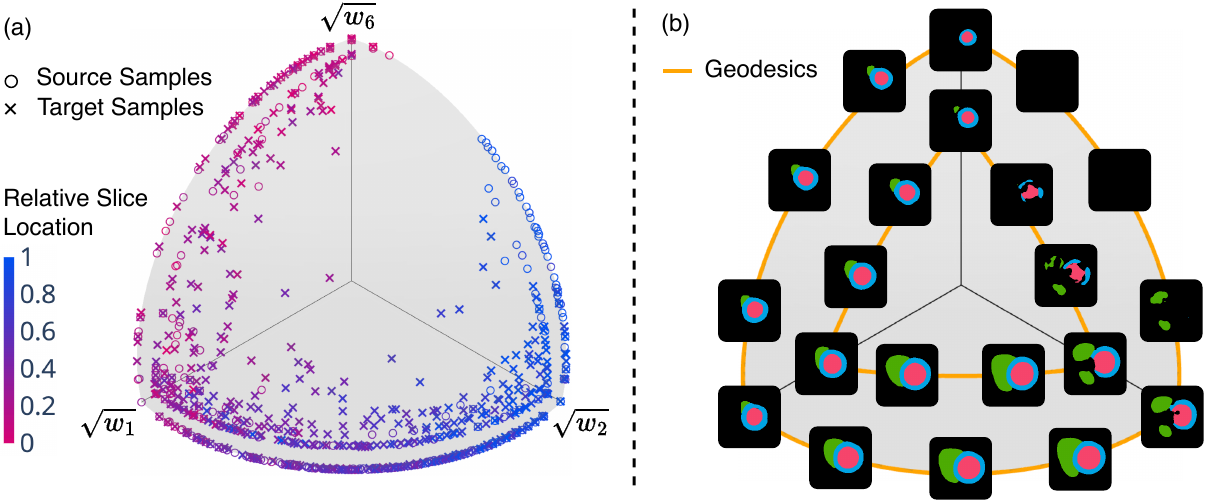}
\caption{3D visualization of the positive orthant of the unit sphere, with the shape blending weights for images in the MS-CMR dataset. (a) Distribution of the transformed weights $\w^\dagger$. Each point is based on a 2D image slice, with colors indicating its relative location among the total number of slices for the corresponding patient. (b) Manipulating $\w^\dagger$ (thus $\z$) along geodesics induces gradual variations in predicted segmentation (before warped by $\st^{-1}$). Endpoints of the shown six geodesics: (1,0,0), (0,1,0), (0,0,1), (0.99,0.1,0.1), (0.1,0.99,0.1), (0.1,0.1,0.99). Note that the segmentations with broken shapes could still be valid, as a few similar ground-truth labels exist.} 
\label{fig:geo_seg}
\end{figure}

\begin{table}[t!]
\caption{Performance (mean/standard deviation) of \ours without certain components on the MS-CMR dataset. $\checkmark$: the loss term was used with optimal strength; $\infty$: spatial transformations were fixed to identity mappings for all images.}
\centering
\scriptsize
\label{tab:ablation}
% \fontsize{0.8em}{9pt}\selectfont
\setlength{\tabcolsep}{0.45em} % for table column spacing
\begin{tabular}{cccccccccccc}
\toprule
$\Lvel$ &$\Lanchor$ & $\Lalign$ & $\Lgm$ & Mean DSC (\%) & Mean ASSD (mm) & Epochs to Converge\\
\midrule
$\infty$ &\checkmark& \checkmark & \checkmark & \ms{60.7}{8.5} & \ms{5.1}{1.8} & 86\\
\checkmark&& \checkmark & \checkmark & \ms{80.6}{6.5} & \ms{1.5}{0.8} & 2055\\
\checkmark&\checkmark & & \checkmark & \ms{81.1}{6.5} & \ms{1.5}{0.9} & 738\\
\checkmark&\checkmark & \checkmark & & \ms{83.0}{5.2} & \ms{1.3}{0.7} & 6619 \\
\checkmark&\checkmark& \checkmark & \checkmark & \ms{83.1}{5.3} & \ms{1.3}{0.6} & 3867\\
\bottomrule
\end{tabular}
\end{table}

\noindentbf{Visualization of shape blending weights.} The low-dimensional nature of the shape blending weights $\w$ allows for more accurate and informative visualization of domain alignment than t-SNE \cite{tSNE} of high-dimensional features commonly used in previous works. In a specific example of \ours trained on the MS-CMR dataset, three components of the weights, $w_3,w_4,w_5$, are consistently zero for all source and target images. 
% In other words, the network relies solely on the remaining components $(w_1,w_2,w_6)$ to capture structural information. 
Given that $M=6$, this sparsity enables visualizing $\w$ in 3D without any information compression, as shown in \cref{fig:geo_seg}. To better reflect the geometry imposed by the Fisher-Rao metric, we transform $(w_1,w_2,w_6)$ into $\w^\dagger\defas (\sqrt{w_1},\sqrt{w_2},\sqrt{w_6})$, which maps the points onto the positive orthant of the unit sphere, $\{\boldsymbol{t}\in\R^3\mid\boldsymbol{t}\succeq \boldsymbol{0},\boldsymbol{t}^\T\boldsymbol{t}={1}\}$, where the geodesic distance (arc length) between any two points is equal to half of their Fisher-Rao metric. In \cref{fig:geo_seg}(a), alignment between the source and target domains is evident, with most points close to the orthant border and only a few outliers in the target domain. Additionally, the point colors, representing slice locations within the corresponding patient, exhibit a smooth and gradual transition across the orthant. This highlights that the learned manifold effectively captures the spatial continuity of anatomical structures across patients in both domains, even though slice location information was not used for training. \cref{fig:geo_seg}(b) further indicates that the manifold efficiently encodes a wide variety of segmentations with smooth transitions along geodesics. The displayed examples cover nearly all possible topological patterns observed in the ground-truth labels. These visualizations underscore the advantages of \ours in developing an explainable and global understanding of anatomical structures, similar to human memory.

\noindentbf{Ablation Study.} The performance of \ours with different components removed is summarized in \cref{tab:ablation}. 
The results demonstrate that spatial transformation is crucial for accurate segmentation, as the weighted average of anchors alone lack the flexibility to fit structures in every image. Besides, both $\Lanchor$ and $\Lalign$ significantly contribute to segmentation accuracy. Notably, even without the domain alignment term $\Lalign$, our model surpasses multiple baseline methods, showcasing the robustness of \ours in generalizing across domains. This strong generalization likely stems from its ability to retain global segmentation information as prior knowledge through the learned manifold. Moreover, while \ours without $\Lgm$ achieves comparable performance to the full version, its slower convergence underscores the benefit of explicitly regularizing the geometry of the learned manifold to enhance training efficiency.

\section{Conclusion and Discussion}
We have introduced a new paradigm for domain-adaptive medical image segmentation that is distinct from all previous works. Our framework, \ours, constrains image features to a latent probabilistic manifold, effectively capturing accurate anatomical information. This design is interpretable, resembling the human process of retrieving the most appropriate segmentation shape from memory. Furthermore, it enables computationally efficient and geometrically faithful domain alignment, outperforming state-of-the-art methods that usually rely on multiple complicated and ad hoc alignment strategies. 

\ours has the potential to address even more challenging scenarios, such as source-free adaptation \cite{SFDA_survey}. Specifically, during source-domain training, anchor parameters and the shape blending weights $\w$ for all source images could be stored and later used to adapt to the target images without requiring access to the source data --- a promising direction for future research. A limitation of our method is that sharing anchors across images may reduce the network's flexibility, potentially leading to slightly lower segmentation performance on the source domain. However, as shown in our experiments, this trade-off greatly improved target-domain performance. Besides, this potential limitation could be addressed by increasing the number of anchors ($M$) and employing more advanced network architectures for the encoders and decoders.

% \begin{table}
% \caption{Table captions should be placed above the
% tables.}\label{tab1}
% \begin{tabular}{|l|l|l|}
% \hline
% Heading level &  Example & Font size and style\\
% \hline
% Title (centered) &  {\Large\bfseries Lecture Notes} & 14 point, bold\\
% 1st-level heading &  {\large\bfseries 1 Introduction} & 12 point, bold\\
% 2nd-level heading & {\bfseries 2.1 Printing Area} & 10 point, bold\\
% 3rd-level heading & {\bfseries Run-in Heading in Bold.} Text follows & 10 point, bold\\
% 4th-level heading & {\itshape Lowest Level Heading.} Text follows & 10 point, italic\\
% \hline
% \end{tabular}
% \end{table}

\ifthenelse{\boolean{usebiblatex}}{
\printbibliography
}{
\bibliographystyle{splncs04}
\bibliography{ref}

\begin{thebibliography}{10}
\providecommand{\url}[1]{\texttt{#1}}
\providecommand{\urlprefix}{URL }
\providecommand{\doi}[1]{https://doi.org/#1}

\bibitem{velocity}
Ashburner, J.: A fast diffeomorphic image registration algorithm. Neuroimage  \textbf{38}(1),  95--113 (2007)

\bibitem{human_vision}
Biederman, I.: Recognition-by-components: A theory of human image understanding. Psychological Review  \textbf{94},  115--147 (1987)

\bibitem{SIFA}
Chen, C., Dou, Q., Chen, H., Qin, J., Heng, P.A.: Unsupervised bidirectional cross-modality adaptation via deeply synergistic image and feature alignment for medical image segmentation. IEEE Transactions on Medical Imaging  \textbf{39}(7),  2494--2505 (2020). \doi{10.1109/TMI.2020.2972701}

\bibitem{ot_uda}
Courty, N., Flamary, R., Tuia, D., Rakotomamonjy, A.: Optimal transport for domain adaptation. IEEE Transactions on Pattern Analysis and Machine Intelligence  \textbf{39}(9),  1853--1865 (2017). \doi{10.1109/TPAMI.2016.2615921}

\bibitem{VAMCEI}
Cui, H., Li, Y., Wang, Y., Xu, D., Wu, L.M., Xia, Y.: Toward accurate cardiac mri segmentation with variational autoencoder-based unsupervised domain adaptation. IEEE Transactions on Medical Imaging  \textbf{43}(8),  2924--2936 (2024). \doi{10.1109/TMI.2024.3382624}

\bibitem{kl_v}
Dalca, A.V., Balakrishnan, G., Guttag, J., Sabuncu, M.R.: Unsupervised learning of probabilistic diffeomorphic registration for images and surfaces. Medical Image Analysis  \textbf{57},  226--236 (2019). \doi{https://doi.org/10.1016/j.media.2019.07.006}, \url{https://www.sciencedirect.com/science/article/pii/S1361841519300635}

\bibitem{sinkhorn}
Feydy, J., S\'{e}journ\'{e}, T., Vialard, F.X., Amari, S.i., Trouve, A., Peyr\'{e}, G.: Interpolating between optimal transport and mmd using sinkhorn divergences. In: Chaudhuri, K., Sugiyama, M. (eds.) Proceedings of the Twenty-Second International Conference on Artificial Intelligence and Statistics. Proceedings of Machine Learning Research, vol.~89, pp. 2681--2690. PMLR (2019), \url{https://proceedings.mlr.press/v89/feydy19a.html}

\bibitem{DA_survey}
Guan, H., Liu, M.: Domain adaptation for medical image analysis: A survey. IEEE Transactions on Biomedical Engineering  \textbf{69}(3),  1173--1185 (2022). \doi{10.1109/TBME.2021.3117407}

\bibitem{AdaIN}
Huang, X., Belongie, S.: Arbitrary style transfer in real-time with adaptive instance normalization. In: 2017 IEEE International Conference on Computer Vision (ICCV). pp. 1510--1519 (2017). \doi{10.1109/ICCV.2017.167}

\bibitem{atlas_survey}
Iglesias, J.E., Sabuncu, M.R.: Multi-atlas segmentation of biomedical images: A survey. Medical Image Analysis  \textbf{24}(1),  205--219 (2015). \doi{https://doi.org/10.1016/j.media.2015.06.012}, \url{https://www.sciencedirect.com/science/article/pii/S1361841515000997}

\bibitem{amos22}
Ji, Y., Bai, H., Yang, J., Ge, C., Zhu, Y., Zhang, R., Li, Z., Zhang, L., Ma, W., Wan, X., et~al.: Amos: A large-scale abdominal multi-organ benchmark for versatile medical image segmentation. arXiv preprint arXiv:2206.08023  (2022)

\bibitem{vae}
Kingma, D.P., Welling, M.: {Auto-Encoding Variational Bayes}. In: 2nd International Conference on Learning Representations, {ICLR} 2014, Banff, AB, Canada, April 14-16, 2014, Conference Track Proceedings (2014)

\bibitem{SFDA_survey}
Liu, Y., Chen, Y., Dai, W., Gou, M., Huang, C.T., Xiong, H.: Source-free domain adaptation with domain generalized pretraining for face anti-spoofing. IEEE Transactions on Pattern Analysis and Machine Intelligence  \textbf{46}(8),  5430--5448 (2024). \doi{10.1109/TPAMI.2024.3370721}

\bibitem{geo_mean}
Lorenzen, P., Prastawa, M., Davis, B., Gerig, G., Bullitt, E., Joshi, S.: Multi-modal image set registration and atlas formation. Medical image analysis  \textbf{10}(3),  440--451 (2006)

\bibitem{bingo_arxiv}
Luo, X., Wang, X., Shapiro, L., Yuan, C., Feng, J., Zhuang, X.: Bayesian unsupervised disentanglement of anatomy and geometry for deep groupwise image registration (2024), \url{https://arxiv.org/abs/2401.02141}

\bibitem{tSNE}
van~der Maaten, L., Hinton, G.: Visualizing data using t-sne. Journal of Machine Learning Research  \textbf{9}(86),  2579--2605 (2008), \url{http://jmlr.org/papers/v9/vandermaaten08a.html}

\bibitem{fisher-rao}
Miyamoto, H.K., Meneghetti, F.C.C., Pinele, J., Costa, S.I.R.: On closed-form expressions for the fisher--rao distance. Information Geometry  (Sep 2024). \doi{10.1007/s41884-024-00143-2}, \url{https://doi.org/10.1007/s41884-024-00143-2}

\bibitem{att_unet}
Oktay, O., Schlemper, J., Folgoc, L.L., Lee, M., Heinrich, M., Misawa, K., Mori, K., McDonagh, S., Hammerla, N.Y., Kainz, B., Glocker, B., Rueckert, D.: Attention u-net: Learning where to look for the pancreas. In: Medical Imaging with Deep Learning (2018), \url{https://openreview.net/forum?id=Skft7cijM}

\bibitem{VQVAE}
van~den Oord, A., Vinyals, O., kavukcuoglu, k.: Neural discrete representation learning. In: Guyon, I., Luxburg, U.V., Bengio, S., Wallach, H., Fergus, R., Vishwanathan, S., Garnett, R. (eds.) Advances in Neural Information Processing Systems. vol.~30. Curran Associates, Inc. (2017), \url{https://proceedings.neurips.cc/paper_files/paper/2017/file/7a98af17e63a0ac09ce2e96d03992fbc-Paper.pdf}

\bibitem{pytorch}
Paszke, A., Gross, S., Massa, F., Lerer, A., Bradbury, J., Chanan, G., Killeen, T., Lin, Z., Gimelshein, N., Antiga, L., Desmaison, A., K\"{o}pf, A., Yang, E., DeVito, Z., Raison, M., Tejani, A., Chilamkurthy, S., Steiner, B., Fang, L., Bai, J., Chintala, S.: PyTorch: an imperative style, high-performance deep learning library. Curran Associates Inc., Red Hook, NY, USA (2019)

\bibitem{unet}
Ronneberger, O., Fischer, P., Brox, T.: U-net: Convolutional networks for biomedical image segmentation. In: Navab, N., Hornegger, J., Wells, W.M., Frangi, A.F. (eds.) Medical Image Computing and Computer-Assisted Intervention -- MICCAI 2015. pp. 234--241. Springer International Publishing, Cham (2015)

\bibitem{nvae}
Vahdat, A., Kautz, J.: Nvae: A deep hierarchical variational autoencoder. In: Larochelle, H., Ranzato, M., Hadsell, R., Balcan, M., Lin, H. (eds.) Advances in Neural Information Processing Systems. vol.~33, pp. 19667--19679. Curran Associates, Inc. (2020), \url{https://proceedings.neurips.cc/paper_files/paper/2020/file/e3b21256183cf7c2c7a66be163579d37-Paper.pdf}

\bibitem{ADVENT}
Vu, T.H., Jain, H., Bucher, M., Cord, M., Pérez, P.: Advent: Adversarial entropy minimization for domain adaptation in semantic segmentation. In: 2019 IEEE/CVF Conference on Computer Vision and Pattern Recognition (CVPR). pp. 2512--2521 (2019). \doi{10.1109/CVPR.2019.00262}

\bibitem{CyCMIS}
Wang, R., Zheng, G.: Cycmis: Cycle-consistent cross-domain medical image segmentation via diverse image augmentation. Medical Image Analysis  \textbf{76},  102328 (2022). \doi{https://doi.org/10.1016/j.media.2021.102328}, \url{https://www.sciencedirect.com/science/article/pii/S136184152100373X}

\bibitem{mri_cal_seg}
Wang, X., Canton, G., Guo, Y., Zhang, K., Akcicek, H., Yaman~Akcicek, E., Hatsukami, T., Zhang, J., Sun, B., Zhao, H., Zhou, Y., Shapiro, L., Mossa-Basha, M., Yuan, C., Balu, N.: Automated mri-based segmentation of intracranial arterial calcification by restricting feature complexity. Magnetic Resonance in Medicine  \textbf{93}(1),  384--396 (2025). \doi{https://doi.org/10.1002/mrm.30283}, \url{https://onlinelibrary.wiley.com/doi/abs/10.1002/mrm.30283}

\bibitem{BInGo}
Wang, X., Luo, X., Zhuang, X.: Bingo: Bayesian intrinsic groupwise registration via explicit hierarchical disentanglement. In: Frangi, A., de~Bruijne, M., Wassermann, D., Navab, N. (eds.) Information Processing in Medical Imaging. pp. 319--331. Springer Nature Switzerland, Cham (2023)

\bibitem{VarDA}
Wu, F., Zhuang, X.: Unsupervised domain adaptation with variational approximation for cardiac segmentation. IEEE Transactions on Medical Imaging  \textbf{40}(12),  3555--3567 (2021). \doi{10.1109/TMI.2021.3090412}

\bibitem{DARUNet}
Yao, K., Su, Z., Huang, K., Yang, X., Sun, J., Hussain, A., Coenen, F.: A novel 3d unsupervised domain adaptation framework for cross-modality medical image segmentation. IEEE Journal of Biomedical and Health Informatics  \textbf{26}(10),  4976--4986 (2022). \doi{10.1109/JBHI.2022.3162118}

\bibitem{MAPSeg}
Zhang, X., Wu, Y., Angelini, E., Li, A., Guo, J., Rasmussen, J.M., O'Connor, T.G., Wadhwa, P.D., Jackowski, A.P., Li, H., Posner, J., Laine, A.F., Wang, Y.: Mapseg: Unified unsupervised domain adaptation for heterogeneous medical image segmentation based on 3d masked autoencoding and pseudo-labeling. In: Proceedings of the IEEE/CVF Conference on Computer Vision and Pattern Recognition (CVPR). pp. 5851--5862 (June 2024)

\bibitem{mscmr}
Zhuang, X., Xu, J., Luo, X., Chen, C., Ouyang, C., Rueckert, D., Campello, V.M., Lekadir, K., Vesal, S., RaviKumar, N., et~al.: Cardiac segmentation on late gadolinium enhancement mri: a benchmark study from multi-sequence cardiac mr segmentation challenge. Medical Image Analysis  \textbf{81},  102528 (2022)

\end{thebibliography}
}

\end{document}